\documentclass[10pt,twocolumn,letterpaper]{article}

\usepackage{cvpr}              

\usepackage{graphicx}
\usepackage{amsmath}
\usepackage{amssymb}
\usepackage{booktabs}
\usepackage{tikz}
\usepackage{comment}
\usepackage{color}
\usepackage{adjustbox}
\usepackage{wrapfig}
\usepackage{multirow}
\usepackage{float}

\usepackage[pagebackref,breaklinks,colorlinks]{hyperref}

\usepackage[capitalize]{cleveref}
\crefname{section}{Sec.}{Secs.}
\Crefname{section}{Section}{Sections}
\Crefname{table}{Table}{Tables}
\crefname{table}{Tab.}{Tabs.}

\def\cvprPaperID{4912} 
\def\confName{CVPR}
\def\confYear{2023}

\begin{document}

\title{StyleTalker: One-shot Style-based Audio-driven Talking Head Video Generation}


\author{Dongchan Min\thanks{ Equal contribution}    , Minyoung Song\footnotemark[1]    , Eunji Ko, Sung Ju Hwang \\ 
Graduate School of AI, Korea Advanced Institute of Science and Technology (KAIST) \\ 
Seoul, South Korea \\ 
{\tt\small \{alsehdcks95, mysong105, kosu7071, sjhwang82\}@kaist.ac.kr} 
}

\maketitle

\begin{abstract}
   We propose \emph{StyleTalker}, a novel audio-driven talking head generation model that can synthesize a video of a talking person from a single reference image with accurately audio-synced lip shapes, realistic head poses, and eye blinks. Specifically, by leveraging a style-based generator, we estimate the latent codes of the talking head video that faithfully reflects the given audio. This is made possible with several newly devised components: 1) A contrastive lip-sync discriminator for accurate lip synchronization, 2) A conditional sequential variational autoencoder that learns the latent motion space disentangled from the lip movements, such that we can independently manipulate the motions and lip movements while preserving the identity. 3) An auto-regressive prior augmented with normalizing flow to learn a complex audio-to-motion multi-modal latent space. Equipped with these components, StyleTalker can generate talking head videos not only in a motion-controllable way when another motion source video is given but also in a completely audio-driven manner by inferring realistic motions from the input audio. Through extensive experiments and user studies, we show that our model is able to synthesize talking head videos with impressive perceptual quality which are accurately lip-synced with the input audios, largely outperforming state-of-the-art baselines.
\end{abstract}

\section{Introduction}
\label{sec:intro}

Recently, bridging the gap between the real and the virtual world with visual media is becoming increasingly important, as the coronavirus pandemic has completely changed the way we work, learn, play, and socialize. This has led to the emergence of \emph{digital twins}, which are digital copies of real human beings that can generate realistic speech and facial expressions to mimic the target person, as they are useful for real-world applications such as virtual conferencing and VR/AR platforms. Audio-driven talking head generation, which aims to synthesize a video of a talking person with the input audio stream for multi-speakers, is a crucial technology for implementing such digital twins.  

\begin{figure*}
    \centering
    \includegraphics[width=0.8\linewidth]{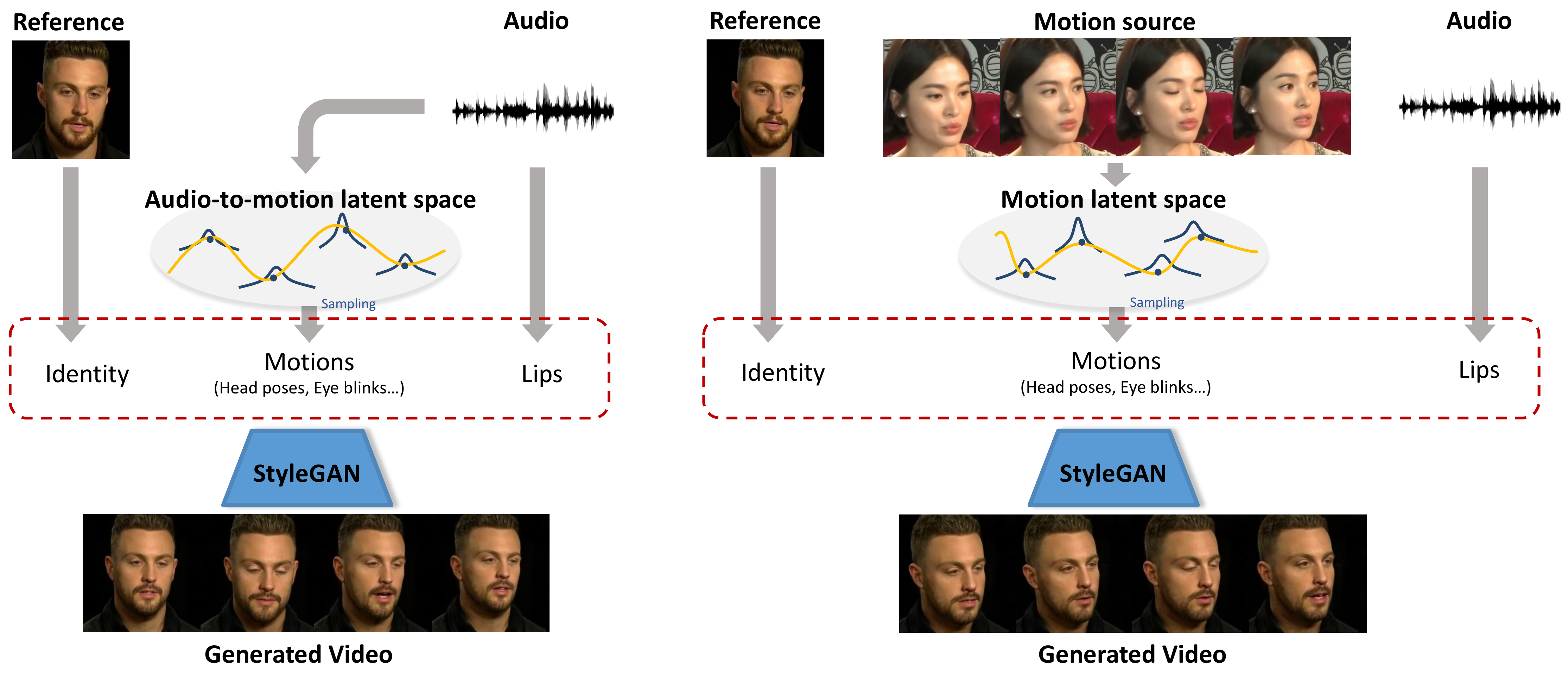}
    \caption{\small \textbf{Concept}. StyleTalker generates a realistic talking head video from a single reference image and input audio in two different ways. \textbf{Left: Audio-driven Generation} Motions are predicted from the audio. \textbf{Right: Motion-controllable Generation} Motions are controlled by another motion source video.}
    \label{fig:intro}
\end{figure*}

While some of the previous works from the graphics area~\cite{obama,nvp,lip3d} have achieved some success in generating realistic digital twins, they target single-speaker generation and thus have no control over multiple identities. One-shot audio-driven talking head generation methods~\cite{wav2lip,MakeltTalk,a2h,pcavs}, on the other hand, aim to generate a talking head video using a single reference image of the target person and arbitrary audio, and have received much attention recently to its applicability in many applications. However, previous one-shot audio-driven talking head generation methods only synthesize around the lip region~\cite{wav2lip} or fail to predict realistic motions (head poses and eye blinks) due to the nature of difficulty in modeling motions from the audio~\cite{MakeltTalk,a2h}. To overcome these issues, Zhou \etal ~\cite{pcavs} recently proposed a pose-controllable method to copy the head poses of a given video independently with the given audio, but its applicability is limited since it requires another video as the pose source and also fails to capture perceptually important facial attributes such as eye blinks. 

In this paper, we propose \emph{StyleTalker}, a novel audio-driven talking head generation model. Our model leverages a style-based generator, namely StyleGAN~\cite{stylegan,stylegan2,stylegan3}, which learns a rich distribution of human faces, and GAN inversion techniques~\cite{psp, e4e} to embed images onto the style latent space. It has been shown that the style latent space accurately represents facial attributes with disentangled properties~\cite{psp,e4e}, allowing extensive image manipulations~\cite{image2stylegan,image2stylegan++}. We can generate a talking head video by sequentially estimating the latent codes of each frames in the video. However, this is extremely challenging, since estimated latent codes should accurately reflect the given audio and motions while keeping the identity, in order to generate a natural and realistic talking head video.




To handle this challenge, we first pretrain our lip-sync discriminator with a novel contrastive learning objective and utilize it as a fixed discriminator during training to improve the lip synchronization with the input audio. Furthermore, we extract motions from the video using a sequential variational autoencoder (VAE)~\cite{vae} to learn the posterior distribution of motion latent space conditioned on the audio. In addition, we design an auto-regressive prior augmented with normalizing flow to capture more complicated probabilistic distributions between audio and possible motions. Then, we manipulate the latent code of the reference image to change the lip shapes, head poses, and eye shapes according to the given audio and motions to reconstruct the video while keeping the identity fixed. In this way, our model can learn their implicit representations without the help of any geometry priors such as landmarks or 3D pose parameters. 

As a result, during inference, our StyleTalker can generate realistic talking head videos either by sampling the motions from the motion latent space or the audio-to-motion latent space. We refer to the former as an \textit{motion-controllable} talking head video generation and the latter as a \textit{audio-driven} talking head video generation. We illustrate the overall concept of StyleTalker in Fig.~\ref{fig:intro}. The following is the summary of our contributions. 

\begin{itemize}
\item We propose StyleTalker, a novel one-shot audio-driven talking head generation framework, that controls lip movements, head poses, and eye blinks using a pretrained image generator by learning their implicit representations in an unsupervised manner.

\item StyleTalker can generate talking head videos either in a motion-controllable manner, using other videos as motion sources, or in a completely audio-driven manner, predicting realistic motions from the audio.

\item StyleTalker achieves state-of-art talking head generation performance, generating more realistic videos with accurate lip syncing and natural motions compared to baselines, both by quantitative metrics and user studies.
\end{itemize}

\section{Related Work}


\begin{figure*}
    \centering
    \includegraphics[width=0.95\linewidth]{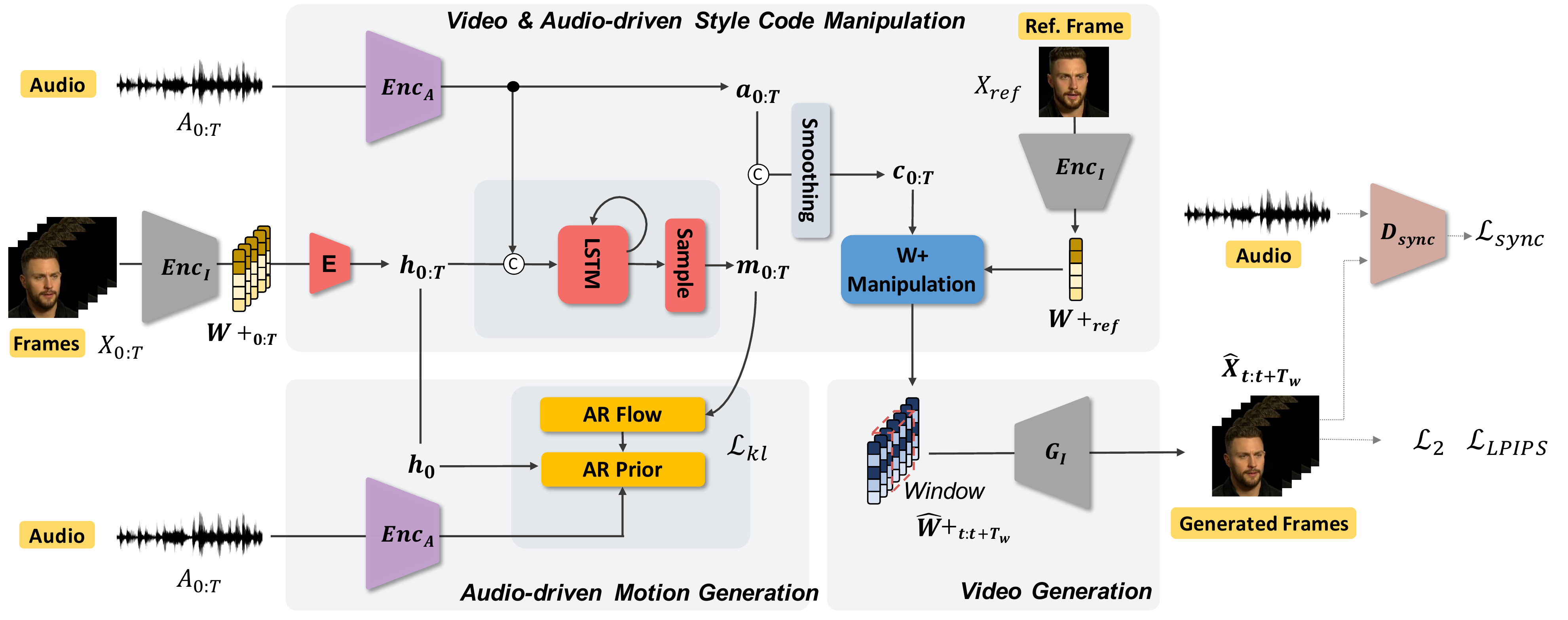}
    \caption{\small \textbf{The overall training framework of StyleTalker.} 1) \textit{Video \& Audio driven Style Code Manipulation}: Given a video, StyleTalker maps audio into audio features $a_{0:T}$ and frames into motion latent variables $m_{0:T}$. With $a_{0:T}$ and $m_{0:T}$, we manipulate the style latent code of the reference frame $\mathbf{\mathcal{W}}+_{ref}$ to generate $\widehat{\mathbf{\mathcal{W}}}+_{t:t+T_w}$. 2)\textit{Video Generation}: $\widehat{\mathbf{\mathcal{W}}}+_{t:t+T_w}$ are forwarded to generate $n$ sequential frames, which are the reconstruction of the given video. 3)\textit{Audio-driven Motion Generation}: We model our prior network of $m_{0:T}$ with given audio and an image as inputs. Note that $Enc_I$, $G_I$, and $D_{sync}$ are fixed during training.}
    \label{fig:TraingOverview}
\end{figure*}
\vspace{-0.1in}

\noindent \textbf{Neural Lip-synced Video generation.}
Generating an audio-synchronized video of a target person has progressed rapidly thanks to the recent advances of image generation methods. Many of these methods~\cite{obama,lip3d,nvp} aim toward transforming the lip region of the person in the target video, generating new videos with the lip shapes that match the input audio. However, their applicability to real-world scenarios is somewhat limited as they mainly focus on a single identity and require a large amount of training data for each identity. For example,~\cite{obama} synthesizes a high-quality video of Barack Obama conditioned on his speech using the lip landmarks but requires long hours of videos from the target person for model training (up to 17 hours). Recently, Wav2lip~\cite{wav2lip} handles in-the-wild data with multiple identities without any geometric priors, but the generated images of the model have low resolutions (96$\times$96). Furthermore, the lip-sync methods can only animate lips without generating head movements that are crucial for realistic talking head generation, producing unnatural videos of fixed or minimal motions~\cite{lipgan,chen2019hierarchical}, and fail to generate perceptually realistic videos on unrefined datasets~\cite{voug,das2020speech}.



\noindent \textbf{Audio-driven Talking Head Generation.}
To generate more plausible talking head videos, audio-driven generation methods~\cite{MakeltTalk,rhythmic,pcavs,a2h,nwt,voug} aim to generate head movements in addition to synchronizing lip movements for given audio. Furthermore, many of them assume an image as the identity source handling multi-speaker generation in an one-shot manner. However, this goal is challenging because its requirements are multifaceted: high-quality image generation, multi-speaker handling, lip synchronization, temporal consistency regularization, and natural generation of diverse human movements. Therefore, most recent methods adopt a reconstruction based pipeline~\cite{zhou19,wav2lip,lipgan,MakeltTalk,pcavs,a2h}. Within the pipeline, pre-estimated geometry priors such as landmarks and 3D pose parameters decouple the head pose, identity, and lip movements from the given images and audio signal. For example, MakeItTalk~\cite{MakeltTalk} predicts landmarks and uses them as intermediate representations between the audio and visual features. Moreover, PC-AVS~\cite{pcavs} uses head pose coordinates to decouple the pose features expressed in Euler angles (yaw, pitch, roll). Audio2Head~\cite{a2h}, a warping-based method, also maximizes the similarity between the head pose coordinates and the predicted head poses. However, such geometry priors are insufficient to describe natural human motions. Thus, we adopt a reconstruction-based pipeline but design a novel framework that does not use any geometric priors.

\section{Method}


Here, we describe our model \textbf{StyleTalker} which synthesizes a talking head video from a single reference image of a target person either in a motion-controllable manner or in a audio-driven manner. Specifically, for training, our model takes a video $V$ with $T$ frames $V=(X_0, X_1, ..., X_T)$, a single reference image of the target person $X_{ref}$, and a sequence of $T$ audio spectrograms $A=(A_0, A_1, ..., A_T)$. Then, we extract motions $m_{0:T}=(m_o, m_1, ..., m_T) $ and audio features $a_{0:T}=(a_0, a_1, ..., a_T) $ from the video $V$ and the sequence of audio spectrograms $A$, respectively. The training objective is to reconstruct the video $V$ by reflecting extracted motions and speech information to the reference image $X_{ref}$ while keeping the identity. Furthermore, we model an audio-to-motion distribution which maps speech information $a_{0:T}$ to motions $m_{0:T}$. At the inference time, we can generate a talking head video with an identity, motions, and speech information which are originated from different sources. In particular, motions can be either extracted from a completely different video (motion-controllable) or inferred from an audio through audio-to-motion distribution (audio-driven).

To this end, we propose several components: 1) A lip-sync discriminator, trained with contrastive learning for precise and natural lip movements, 2) The conditional sequential VAE which learn the posterior distribution of
motion latent space conditioned on the audio, 3) Auto-regressive prior augmented with normalizing flow for learning complex audio-motion latent space, 4) Style code manipulation to estimate the style codes of the target video. The overall training procedure of StyleTalker is illustrated in Fig~\ref{fig:TraingOverview}.

\subsection{Contrastive Lip-sync Discriminator}
\label{section:syncnet}

\begin{figure}[t]
  \centering
  \includegraphics[width=0.8\linewidth]{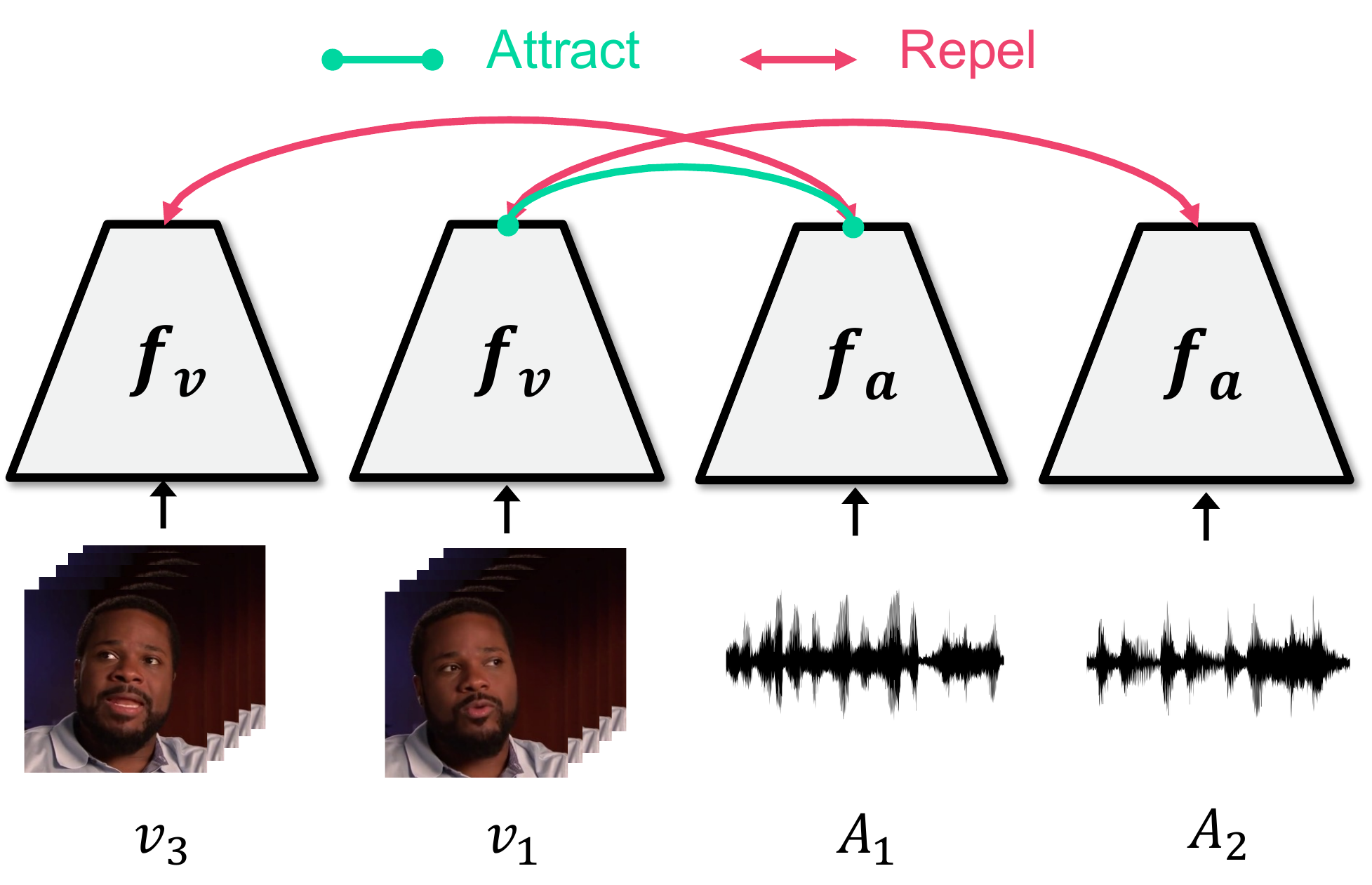}
  \caption{\small \textbf{Contrastive Lip-sync Discriminator}}
  \label{fig:SyncNet}
\end{figure}

Recently, Prajwal \etal~\cite{wav2lip} demonstrated that a strong lip sync expert plays a very important role in generating precise and natural lip movements. Similarly, we also use a lip-sync discriminator, $D_{sync}$, which predicts whether the given audio-video pair is synced or not. However, instead of modeling it as a binary classification problem as Prajwal \etal did, we use contrastive learning to make in-sync pairs align closer than off-sync audio-video pairs, as shown in Fig~\ref{fig:SyncNet}. Contrastive learning is known to maximize a lower bound of the mutual information between two modalities~\cite{InfoNCE}, and has largely improved the performance of several recent generative models~\cite{xmcgan,contragan}. 
Specifically, given a window $V_k = X_{k-T_s:k+T_{s+1}}$ with $2\cdot T_s+1$ consecutive frames and an audio segment $A_{k}$, the score function is as follows: 
\begin{equation}
    S_{sync} = cos(f_v(V_k), f_a(A_{k})) / \tau
\end{equation}
where $cos(\cdot,\cdot)$ denotes the cosine similarity, $\tau$ denotes a temperature hyper-parameter, and $f_v$ and $f_a$ are video and audio encoders to extract video and audio feature vectors onto a joint embedding space. The contrastive loss between a video $V_k$ and its in-sync paired audio segment $A_k$ is then computed as:
\begin{equation}
    \mathcal{L}_{c} = -\log \, \frac{cos(f_v(V_k), f_a(A_{k})) / \tau}
    {\sum^M_{i=1} cos(f_v(V_i), f_a(A_{i})) / \tau}
\end{equation}
where $M$ is the total number of video-audio pairs. The off-sync pairs can be sampled from the same video by shifting each video frame along the time axis while fixing the audio segments and vise versa. Following \cite{wav2lip}, we utilize the pretrained lip-sync discriminator as a fixed discriminator to guide the lip movements of the generated videos.


\subsection{GAN Inversion}
\label{section:image_encoding}
For StyleGAN~\cite{stylegan,stylegan2,stylegan3}, several GAN inversion techniques have proposed image encoder networks which embed the images onto style latent space, which is known as $\mathcal{W}+$ space~\cite{psp,e4e}. In this work, we embed video frames onto style latent space using pixel2style2pixel (pSp)~\cite{psp} encoder to deposit facial attributes such as head poses, eye shapes and lip shapes. In detail, the image encoder $Enc_I$ outputs a latent style vector as $\mathcal{W}+ = Enc_I(X)$, where $X \in \mathbb{R}^{3\times H \times W}$ is an input frame and $\mathcal{W}+ \in \mathbb{R}^{L \times 512}$ consists of $L$ style codes, $w^i \in \mathbb{R}^{512}$, that are fed into each layer of a StyleGAN with $L$ layers. Then, we independently process each style code, $w^i$, which enables StyleTalker to perform delicate transformations of the facial attributes to generate frames in the video. For readability, we omit $i$ in the rest of the paper.





\subsection{Conditional Sequential VAE}

Let $w_{ref}$ denote a style code of the reference image. Further, Let $w_{0:T} = (w_0, w_1, ..., w_T)$ and $ a_{0:T} = (a_0, a_1, ..., a_T)$ denote $T$ style codes of a video with $T$ consecutive frames and a sequence of $T$ audio features respectively: $w_{0:T}=Enc_I(X_{0:T}), \, a_{0:T}=Enc_A(A_{0:T})$. We propose a conditional sequential variational autoencoder model, where the sequence $w_{0:T}$ is generated from $w_{ref}$, the audio features $a_{0:T}$ and latent variables $m_{0:T}$, each of which contains identity, lip, and motion information respectively.

Then, we use variational inference to approximate the posterior distribution of $m_t$:
\begin{equation}
    m_t \sim N(\mu_t, diag(\sigma_t))
\end{equation}
More specifically, since $m_t$ is a time-varying variable, we estimate it conditioned on the current and previous frames $w_{\leq t}$, and the audio features $a_{\leq t}$. The posterior distribution of $m_t$ is then defined as follows:
\begin{gather} 
    [\mu_t, \sigma_t] =\psi_R(w_{\leq t}, a_{\leq t})\\
    q_{\phi}(m_{0:T}|w_{0:T}, a_{0:T}) = \prod^{T}_{t=0} q_{\phi}(m_t| w_{\leq t}, a_{\leq t})
\end{gather}
where $\psi_R$ can be modeled as a recurrent network, such as LSTM~\cite{lstm} or GRU~\cite{gru}.

\noindent \textbf{Auto-regressive prior with normalizing flow.}
We assume that the motion at current time step depends on the previous motions and audio features. Thus, we design an prior of $m_t$ as auto-regressive model using a recurrent neural network $\phi_R$. Furthermore, for time step 0, we formulate the prior of $m_0$ conditioned on the first frame $w_0$, since without conditioning, the initial motion may have a large variance, which may prevent the learning of a proper prior distribution in consecutive frames. Specifically, we define the prior distribution of $m_t$ as follows:
\begin{gather}
    m_t \sim N(\mu_t, diag(\sigma_t))\\
    [\mu_t, \sigma_t] =\phi_R(m_{< t}, a_{\leq t}),  \,\, t>0 \\
    [\mu_0, \sigma_0] =\phi_0(w_0)\\
 p_{\theta}(m_{0:T}|a_{1:T}, w_0) = p_{\theta}(m_0|w_0) \prod^{T}_{t=1} p_{\theta}(m_t|m_{< t}, a_{\leq t}, w_0)
\end{gather}
Note that, at the inference time, $w_0$ can be replaced with $w_{ref}$. That is, the generated video could be initialized with the motion of the given reference image.

However, learning an audio-to-motion multi-modal distribution is very challenging, as previous works~\cite{MakeltTalk,a2h} fail to synthesize realistic motions from the audio. We also found that our auto-regressive prior encoder is not sufficient for modeling realistic and natural motions from the audio, so that we need a more complicated distribution for prior than a diagonal Gaussian distribution. To this end, we apply auto-regressive normalizing flow, $f_\theta$, consisting of invertible non-linear transformations with simple Jacobian determinants as follows~\cite{VIwithNF,Flowtron} :
\begin{equation}
    p'_{\theta}(m_{0:T}|a_{1:T}, w_0) = p_{\theta}(m_{0:T}|a_{1:T}, w_0) \Big| det\frac{\delta f_\theta}{\delta m}\Big|
\end{equation}
We then define the KL-divergence loss between the posterior and the prior to train the conditional sequential VAE:
\begin{multline}
    L_{kl} = \log(q_{\phi}(m_{0:T}|a_{0:T}, w_{0:T}))\\ - \log(p'_{\theta}(m_{0:T}|a_{1:T}, w_0))
\end{multline}


\begin{figure}[t]
  \centering
  \includegraphics[width=0.8\linewidth]{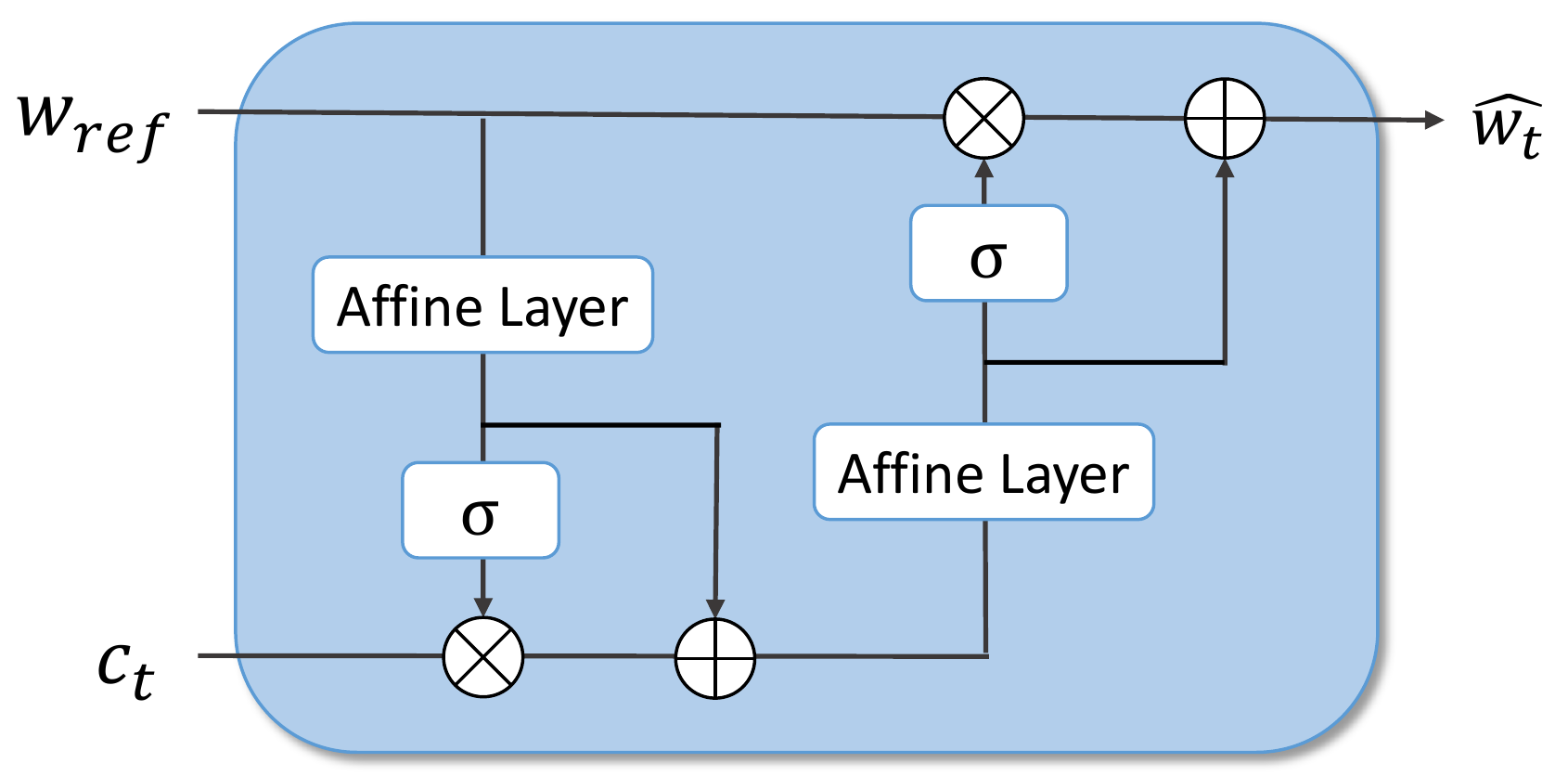}
  \caption{\small \textbf{$\bf w$ manipulation module.}}
  \label{fig:WManipulation}
\end{figure}

\label{sec:stylecodemainpulation}
\noindent \textbf{Style Code Manipulation.} After we obtain audio features $a_{0:T}$ and motion latent variables $m_{0:T}$, we use them to generate the style codes $\widehat{w}_{0:T}$ of the target frames from $w_{ref}$, the style code of the reference image. However, we noticed flicking artifacts in the generated videos since $m_{0:T}$ are sampled from a multivariate Gaussian distribution, which induces noise in the time domain. To handle this issue, we introduce a smoothing layer which consists of 1D Gaussian smoothing, 1D convolution, and activation function. Specifically, the smoothing layer receives the concatenated $a_{0:T}$ and $m_{0:T}$ and outputs smoothed vectors $c_{0:T}$.
\begin{equation}
    \label{eq:12}
    c_{0:T} = f(\text{Conv1D}(\text{GaussianSmooth1D}([a_{0:T}, m_{0:T}]))
\end{equation}

\noindent where $f$ is a non-linear activation function. We then manipulate $w_{ref}$ depending on the smoothed latent code $c_{0:T}$. 
As shown in Fig \ref{fig:WManipulation}, this is done in a two-way fashion. Specifically, the style code sequence $\widehat{w}_{0:T}$ corresponding to the output videos is computed as follows:
\begin{equation}
\begin{split}
    & c'_t = \sigma(K_1(w_{ref})) \times c_t + K_2(w_{ref}) \\
    & \widehat{w}_{t} = \sigma(K_3(c'_t)) \times w_{ref} + K_4(c'_t)
\end{split}
\end{equation}

This two-way manipulation is inspired by the following motivations. We first modify the $c_t$ based on $w_{ref}$ to reflect lip shape and motion of the reference frame. Then, we modify the $w_{ref}$ to obtain a target style vector, $\widehat{w}_{t}$, with the correct lip shape and motion. After we generate style vectors, $\widehat{w}_{0:T}$, they are fed into the  pretrained image generator to synthesize the target frames. 


\begin{table*}
\small
\centering
\caption{\textbf{The quantitative results on VoxCeleb2~\cite{vox2}.} Lower the better for LMD and LMD$_{m}$, and higher the better for other metrics.}
\label{tab:vox2_result}
\begin{adjustbox}{max width=\linewidth}
\centering
\begin{tabular}{@{}c|ccccc|ccccc@{}}
\toprule
\multirow{2}{*}{Models}          & \multicolumn{5}{c|}{Seen}                                 & \multicolumn{5}{c}{Unseen} \\ \cmidrule(l){2-11} 
& \bf SSIM$\uparrow$ & \bf MS-SSIM$\uparrow$ & \bf LMD$\downarrow$ & \bf LMD$_{m}$$\downarrow$ & \bf LSE-C $\uparrow$ 
& \bf SSIM$\uparrow$ & \bf MS-SSIM$\uparrow$ & \bf LMD$\downarrow$ & \bf LMD$_{m}$$\downarrow$ & \bf LSE-C $\uparrow$  \\ 

\midrule
\midrule
 
Wav2Lip~\cite{wav2lip}                      
&   0.93 &    0.97  &    1.60   &  1.49     &   7.44
&   0.93   &     0.97  &  1.49     &  1.47     &  7.40  \\

MakeItTalk~\cite{MakeltTalk}                      
&   0.54    &   0.49   &   3.66    &   1.93    &   3.52
&   0.53    &    0.46   &  4.01   &   1.92    &  4.09  \\

Audio2Head~\cite{a2h}                     
&   0.42    &    0.24   &   5.20  &     2.71  &   5.31
&     0.45      &     0.33      &  5.31    &   2.76  &  5.91  \\

PC-AVS~\cite{pcavs}                    
&    0.62   &  0.69     &    3.63   &  2.32     &   7.87
&     0.61 &    0.67   &     3.68  &    2.34   &   8.05 \\

\midrule
StyleTalker (ours)                    
&    \bf 0.62  &    \bf 0.72   &     \bf 2.81  &   \bf 1.84    &  \bf 8.12
&     \bf 0.62  &    \bf 0.71  &    \bf 2.95  &   \bf 1.87    &   \bf 8.53 \\

\end{tabular}
\end{adjustbox}
\end{table*}


\subsection{Training}
\noindent \textbf{Windowed Frame Generation.} Training StyleTalker to generate all frames of a video may require a prohibitively excessive amount of memory. We thus propose to randomly extract a window of $T_w$ consecutive frames, which we refer to as the training window. StyleGAN then takes $\widehat{w}_{t:t+T_w}$ and synthesizes frames only for this window. 
\begin{equation}
\widehat{X}_{t:t+T_w} = G(\widehat{w}_{t:t+T_w})
\end{equation}


\noindent \textbf{Final Objectives.} We use L2 loss and Learned Perceptual Image Path Similarity (LPIPS) loss~\cite{lpips} for generating video frames that match the ground-truth frames as follows:

\begin{equation}
    L_{2} = ||X - \widehat{X}||_2
\end{equation}

\begin{equation}
    L_{LPIPS} = || P(X) - P(\widehat{X}) ||_2
\end{equation}

where $\widehat{X}$ denotes reconstructed frame, $X$ denotes ground-truth frame, $P(\cdot)$ denotes the perceptual feature extractor. Similar to Siarohin \etal~\cite{fomm}, we compute the perceptual loss in multi-resolution by repeatedly downsampling the frame 3 times.


In addition, for synthesizing lip movements synchronized with audio inputs, we apply the sync loss to maximize the cosine similarity between the features from the generated videos and corresponding audios using lip-sync discriminator (Section~\ref{section:syncnet}).
\begin{equation}
    L_{sync} = 1 - cos(f_v(V_k), f_a(A_k))
\end{equation}
Lastly, our overall objective for training StyleTalker is 
\begin{equation}
    L_{total} =  L_{LPIPS} + \lambda_1 L_{2} + \lambda_2 L_{kl} + \lambda_4 L_{sync}
\end{equation}
We set $\lambda_1 = \lambda_2 = \lambda_3 = \lambda_4 = 0.1$ in our experiments.

\subsection{Inference}

Once StyleTalker has been trained, it can generate the talking head videos in two different ways: motion-controllable and audio-driven manner, given audio features $a_{0:T} = Enc_A(A_{0:T})$ and a style vector $w_{ref} = Enc_I(X_{ref})$ of the target person.

\noindent 1) Motion-controllable generation: when the motion source video, $M_{0:T}$, is given, we can sample the motion latent variables, $m_{0:T}$, from the variational posterior $q_\phi$:
\begin{equation} 
    w_{0:T} = Enc_I(M_{0:T}), ~~
    m_{0:T} \sim q_\phi(m_{0:T}|w_{0:T}, a_{0:T})
\end{equation}
2) Audio-driven generation: if the motion source is not available, we can sample the motion latent variables, $m_{0:T}$, from the prior $p_\theta$:
\begin{equation}
    m_{0:T} \sim p'_{\theta}(m_{0:T}|w_{ref}, a_{1:T})
\end{equation}
After the motion latent variables $m_{0:T}$ are sampled from either the posterior distribution or the prior distribution, we can manipulate the $w_{ref}$ with $a_{0:T}$ and $m_{0:T}$ to generate $\widehat{w}_{0:T}$. Lastly, the pretrained image generator can synthesize a video of $T$ image frames from the style vector sequence $\widehat{w}_{0:T}$, which contains accurate lip shapes according to the given audio segments and realistic motions.

\section{Experiments}

\begin{figure*}
    \centering
    \includegraphics[width=0.8\linewidth]{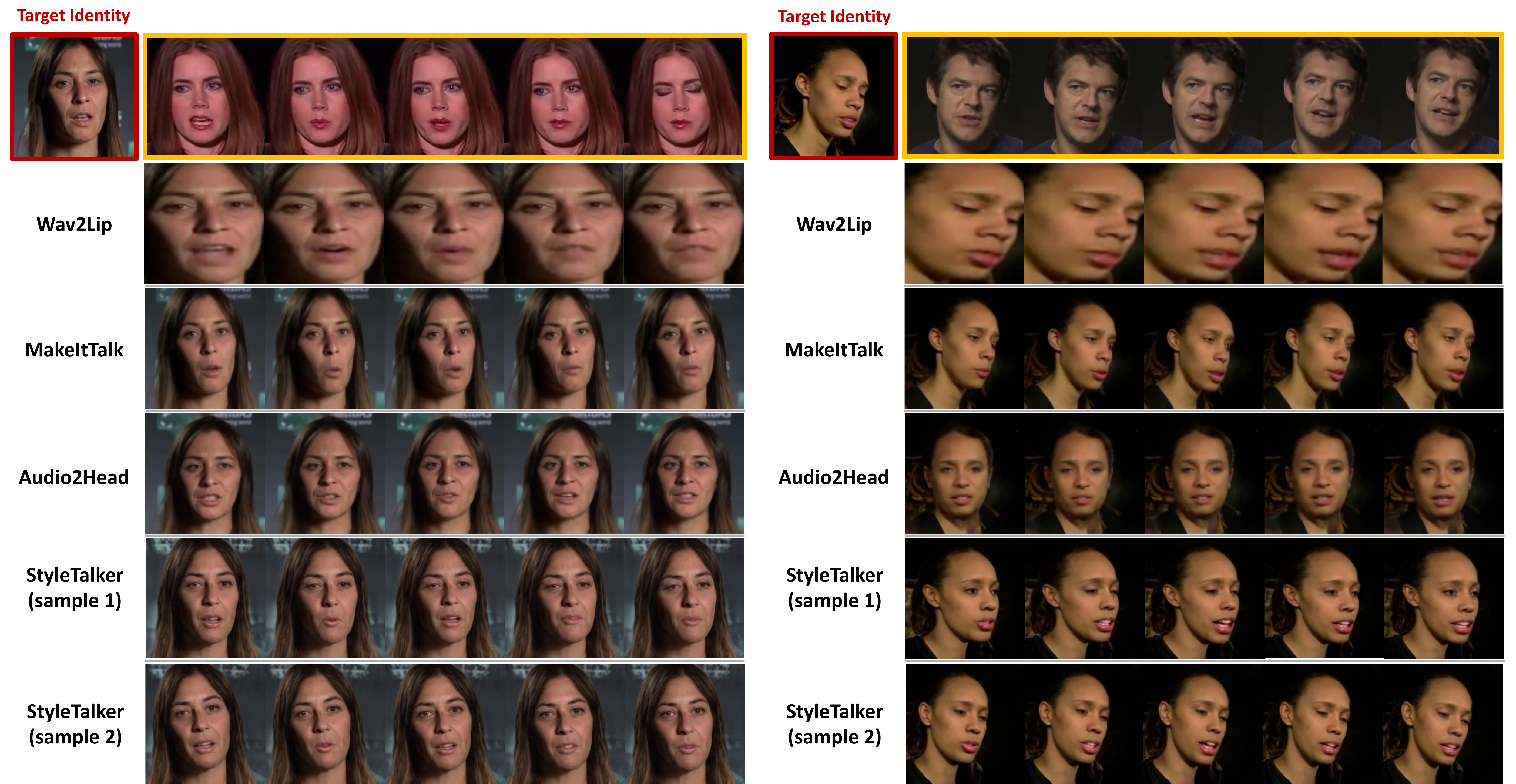}
    \caption{\small \textbf{The qualitative comparison of the audio-driven talking head generation performance on VoxCeleb2.} The first row (yellow box) shows the frames corresponding to the given audio. The single image input (red box) is a reference image of the target identity. 
    \textbf{The Supplementary Materials contain the actual videos.}}
    \label{fig:vox2-audiodriven}
\end{figure*}

\begin{figure*}
    \centering
    \includegraphics[width=0.8\linewidth]{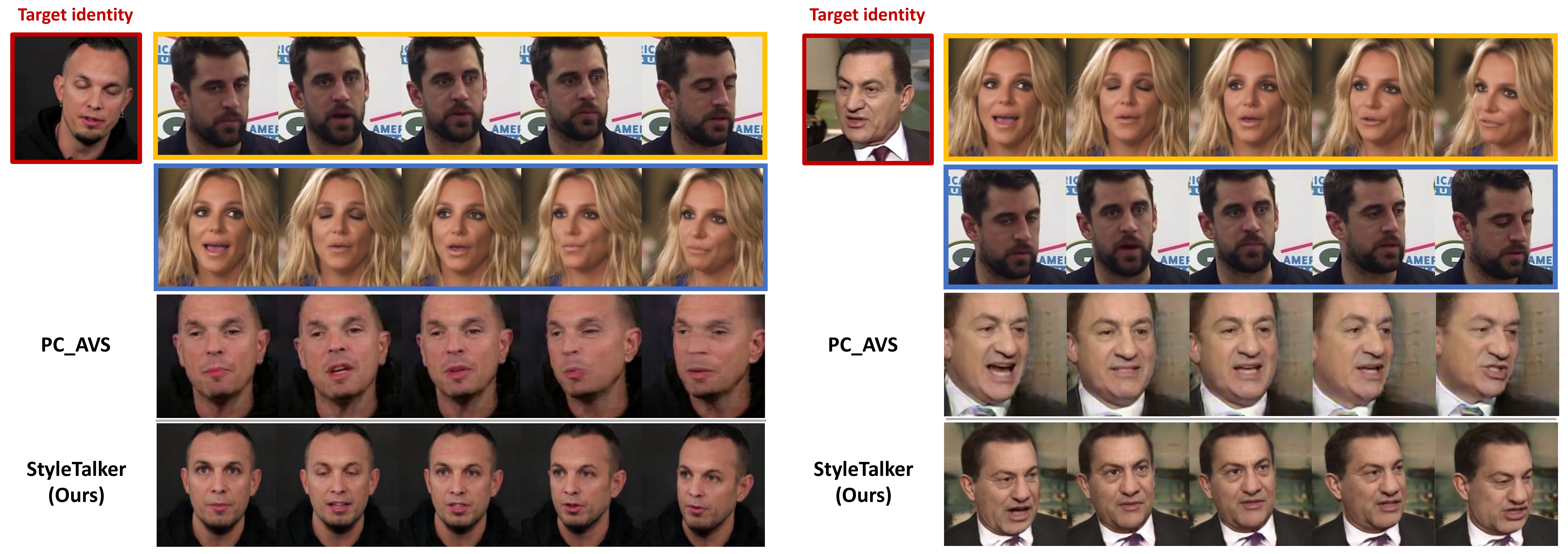}
    \caption{\small \textbf{The qualitative comparison of the motion-controllable generation performance on VoxCeleb2.} The first row (yellow box) shows the frames corresponding to the given audio. The second row (blue box) shows the frames from the motion source video. The single image input (red box) is a reference image of the target identity. \textbf{Please refer to the Supplementary Materials for the actual videos.}}
    \label{fig:vox2-posedriven}
\end{figure*}

\label{dataset}
\noindent \textbf{Dataset.}
We use Voxceleb2~\cite{vox2} dataset collected from the wild for training and test, which is a popularly used dataset in previous works~\cite{pcavs,MakeltTalk,rhythmic}. Voxceleb2 is a dataset which consists of YouTube videos of talking people with a large variety of identities and motions. In detail, it contains 215,000 videos of 6,112 identities over 1 million utterances. We process the video into frames with 25 fps. Following ~\cite{syncnet}, we crop and resize frames into facial-central frames with the size of $256 \times 256$. Audios are down-sampled to 16kHz, from which we extract mel-spectrograms with an FFT size of 512, hop size of 160, a window size of 400 samples, and 80 frequency bins.\\

\noindent \textbf{Implementation Details.}
We pretrain StyleGAN3~\cite{stylegan3} and pSp encoder~\cite{psp} on the Voxceleb2 dataset. Note that these models are kept fixed when training StyleTalker. The style latent space consists of $L=16 $ different style codes and we only manipulate the first 8 style codes which correspond to the coarse and middle level details of the generated images. For each input video frame, we use a 0.2 second long mel-spectrogram. As an audio encoder, we use ResNetSE18~\cite{resnetse}. During training, we generate the style codes with the length of 128 while using the window of 15 style codes to be fed into the StyleGAN3 to generate frames. For training contrastive lip-sync discriminator, we use 5 generated consecutive frames as inputs. We use the Adam optimizer with the learning rate of $1e-4$ for all networks. For more details, including the network architecture specifications and other hyperparameters, please refer to the Supplementary Materials.\\

\begin{table*}
\small
\centering
\caption{\small \textbf{User study reporting MOS with 95\% confidence intervals.} The higher the better for all of the metrics. The number ranges from 1 to 5.}
\label{tab:userstudy}
\begin{adjustbox}{max width=\linewidth}

\begin{tabular}{@{}llc|c|c@{}}
\toprule
\multirow{2}{*}{\bf Methods }                           & \multirow{2}{*}{\bf Models}            & \bf Lip Sync & \bf Motion  &  \bf Video  \\ 
& & \bf Accuracy &\bf Naturalness &\bf Realness \\
\midrule\midrule
\multirow{4}{*}{Audio-driven}        & Wav2Lip~\cite{wav2lip}            
                                    &         3.5 $\pm$ 0.32  &     1.75 $\pm$ 0.31  &     1.61 $\pm$ 0.21   \\
                                  & MakeItTalk~\cite{MakeltTalk}        
                                  &      2.77 $\pm$ 0.37     &     2.52 $\pm$ 0.32  &      2.52 $\pm$ 0.33          \\
                                  & Audio2Head~\cite{a2h}         
                                  &     2.56 $\pm$ 0.38    &     2.68 $\pm$ 0.32    &       2.45 $\pm$ 0.33         \\
                                  & StyleTalker (ours) 
                                  &     \bf 4.06 $\pm$ 0.22    &     \bf  3.47 $\pm$ 0.31  &   \bf  3.11 $\pm$ 0.35        \\
                                   
                                  \midrule
\multirow{2}{*}{Motion-controllable} & PC-AVS~\cite{pcavs}          
                                &            3.70 $\pm$ 0.30       &     2.93 $\pm$ 0.31    &       2.75 $\pm$ 0.36         \\
                                  & StyleTalker (ours) 
                                  &    \bf 3.77 $\pm$ 0.26              &   \bf 3.65 $\pm$ 0.31      & \bf   3.5 $\pm$ 0.35            \\ 
\bottomrule
\end{tabular}

\end{adjustbox}
\end{table*}

\begin{table}[t]
\small
\centering
\caption{\small \textbf{Ablation study.}}
\label{tab:ablation}
\begin{adjustbox}{max width=\linewidth}
\begin{tabular}{@{}lccc@{}}
\toprule
\small
& \multicolumn{1}{c}{$\:$\bf LSE-C$\uparrow$$\:$} & \multicolumn{1}{c}{\begin{tabular}[c]{@{}c@{}} 
\bf Motion\\ \bf Naturalness\end{tabular}} &
\multicolumn{1}{c}{\begin{tabular}[c]{@{}c@{}} 
\bf $\:$ Video\\ $\:$ \bf Realness\end{tabular}} 
\\
\midrule \midrule
StyleTalker &          8.53             &   86.11\%  &  86.11\% \\
w/o Flow    &           8.48            &   13.89\%. &  13.89\% \\
w/o $D_{sync}$ &          6.71             &  N/A & N/A  \\
\bottomrule
\end{tabular}
\end{adjustbox}

\end{table}

\noindent \textbf{Baselines.}
\label{sec:4.2.1}
We compare our method against the following state-of-the-art talking head generation baselines: 1) Wav2Lip~\cite{wav2lip} is a lip-syncing model for videos in the wild. Wav2Lip generates the lower half of the face given the upper half of the face and a target audio. 2) MakeItTalk~\cite{MakeltTalk} is a audio-driven full-head generation model that drives lip movements and motions from audio with 3D facial landmarks. 3) Audio2Head~\cite{a2h} is a audio-driven full-head generation model that utilizes key point based dense motion field to warp the face images using the audio.
4) PC-AVS~\cite{pcavs} is a video-driven full-head generation model that enables pose-controllable talking head generation using the other video as a pose source.


\subsection{Quantitative Results}
\noindent \textbf{Evaluation Metric.} 
To evaluate the quality of the generated talking head videos using our model, we adopt some of the metrics that have been used for the same tasks~\cite{pcavs,MakeltTalk}. Specifically, 
SSIM, MS-SSIM are used to evaluate the quality of the generated frame. For lip sync accuracy, we use the SyncNet~\cite{syncnet} confidence score (LSE-C) metric and LMD$_m$, the distance to the landmark around the mouth. In addition, LMD is the distance of all landmarks on the face to evaluate motion accuracy.

\noindent \textbf{Evaluation Results.} We first measure the reconstruction performance of our model using the ground-truth videos, on videos whose target motions and target audios are unseen during the training time. 
Table~\ref{tab:vox2_result} contains the results for identities which are included in the training set and for new identities which are not included in the training set. The results show that StyleTalker comprehensively outperforms all baselines. As for the lip-sync accuracy, our model obtains superior results by the LMD$_m$ score and achieves the best performance by LSE-C. Furthermore, we achieve better LMD score than baselines which indicates StyleTalker accurately follows the target motions. Note that Wav2Lip shows the highest score because they generate only the lip region at the lowest resolution; however, the method generates perceptually implausible videos since the other parts of the image remains the same as the input frame. This results show that our model can generate precise talking head videos videos from the given audio and motions. 



\subsection{Qualitative Results}
\noindent \textbf{Audio-driven generation.} In Fig~\ref{fig:vox2-audiodriven}, we show the generated video frames by our StyleTalker in comparison to those generated by other audio-driven generation models~\cite{wav2lip,MakeltTalk,a2h}. We see that StyleTalker generates high-quality talking heads which preserve the identity of the target person, while Audio2Head~\cite{a2h} generates ones that have inconsistent identities that do not match the target. Furthermore, we observe that Audio2Head always frontalizes the faces regardless of the initial pose of the reference image, which results in generating unnatural videos. Since our model samples motions from the prior that is conditioned on the initial pose of the reference image, it can generate more natural and robust talking head videos compared to other models. In addition, our model synthesizes accurate lip shapes which match the ground-truth lip shapes, while Audio2Head~\cite{a2h} and MakeItTalk~\cite{MakeltTalk} generate inaccurate lip shapes.
Our model can also synthesize eye blinks while most others~\cite{MakeltTalk,wav2lip,pcavs} fail to do so, which also crucially affects the perceptual quality of the generated videos. We attribute this to our proposed learnable prior with normalizing flow, which learns a rich multi-modal distribution that maps audio to natural and realistic motions. Furthermore, since we model the audio-motion space as probabilistic distributions, we can generate multiple videos with different motions by sampling several times from the same input audio.

\noindent \textbf{Motion-controllable generation.} Additionally, we show the video frames generated by our model and PC-AVS in a motion-controllable manner, in Fig~\ref{fig:vox2-posedriven}. Our model can generate frames with the same motions as the motion source, synthesizing accurate lip movements that match the given audio. This demonstrates that our model successfully disentangles motions and lip movements. PC-AVS~\cite{pcavs} also seems to disentangle pose and lip movements, but the faces generated with the method have unnatural eyes and have less pose variations than those generated with our method.


\vspace{0.07in}
\noindent \textbf{User Study.}  We conduct an user study for the perceptual quality of the videos generated using our model and the baselines. Specifically, we use MOS (mean opinion score) for evaluating the generated videos based on three criteria: lip sync accuracy, motion naturalness, and video realness. MOS are rated by the 1-to-5 scale and reported with the  95\%  confidence  intervals (CI). 20 judges participated in the study, where each evaluated 4 videos generated by ours and other models, presented in random orders. Please see the \emph{Supplementary File} for more details.
Table~\ref{tab:userstudy} shows the results of the user study. Our model outperforms all baselines in three metrics, on both audio-driven and motion-controllable talking face generation tasks, demonstrating that it can generate accurate and natural talking head videos. Wav2Lip~\cite{wav2lip} scores high on audio-visual sync performance, but its motion naturalness and video realness scores are low since it only synthesizes the lower half of the faces. For PC-AVS~\cite{pcavs}, it generates more natural videos than audio-driven methods since it uses real videos as the pose source. However, it fails to synthesis eye blinking, which is important for the perceptual quality of the generated videos. Thus, its motion naturalness and video realness scores are lower than those from our model.


\subsection{Ablation Study}
\label{section:ablation}
We further conduct an ablation study to verify the effectiveness of each component of our model, namely normalizing flow and the lip-sync discriminator. We use three metrics: LSE-C, motion naturalness, and video realness. For motion naturalness and video realness, 12 judges received 6 sets of two videos in a random order, each created with two different methods (w/ and w/o normalizing flow), and were asked to choose the better one. The results of the ablation study are shown in Table~\ref{tab:ablation}. First of all, we observe that most of the judges prefer the videos generated by our model to the ones generated by the model without normalizing flow in terms of motion naturalness and video realness. Specifically, removing normalizing flow resulted in fluctuated motions, and the model even failed to learn eye blinks. Secondly, without the lip-sync discriminator, the LSE-C score dropped, indicating the importance of the lip-sync discriminator for synthesizing accurate lip shapes.



\section{Conclusion}
We propose StyleTalker, a novel framework that can generate natural and realistic videos of human talking heads from a single reference image with diverse motions and lip shapes that are accurately synchronized with the input audio. Specifically, we leverage the power of the style-based generator and estimate the latent style codes to generate a talking head video. To this end, we propose a novel conditional sequential VAE framework augmented with the normalizing flow for modeling complex audio-to-motion distribution. This allows our model to generate realistic talking head videos either in a motion-controllable manner using another video as a motion source, or in a completely audio-driven manner. The experimental results and user study show that StyleTalker achieves state-of-art performance, generating significantly more realistic videos in comparison to baselines.



{\small
\bibliographystyle{ieee_fullname}
\bibliography{egbib}
}

\appendix
\clearpage
\onecolumn

\begin{center}
{\bf \Large{StyleTalker: One-shot Style-based\\Audio-driven Talking Head Video Generation}}
\end{center}
\begin{center}{\bf {Supplementary Material} \linebreak }
\end{center}

Please check \textcolor{red}{\bf `ShowSamples.html'} in the supplementary file, which includes the videos generated by StyleTalker and other baselines.

\section{Detailed model architecture}

\subsection{Overall architecture}
We use StyleGAN3~\cite{stylegan3} as the image generator, $G_I$, and pSp encoder~\cite{psp} as the image encoder, $Enc_I$. We use ResNetSE18~\cite{resnetse} as the audio encoder, $Enc_A$. Among 16 different style codes in the latent style vector, $\mathbf{\mathcal{W}}+_{ref}$, of the reference image, we process first 8 style codes, $w_0 \sim w_7$, independently using each corresponding Conditional Sequential VAE block. Each Conditional Sequential VAE block consists of the posterior encoder, smoothing module, $w$ manipulation module, prior encoder and normalizing flow.

\begin{figure}[htbp]
    \centering
    \includegraphics[width=0.9\linewidth]{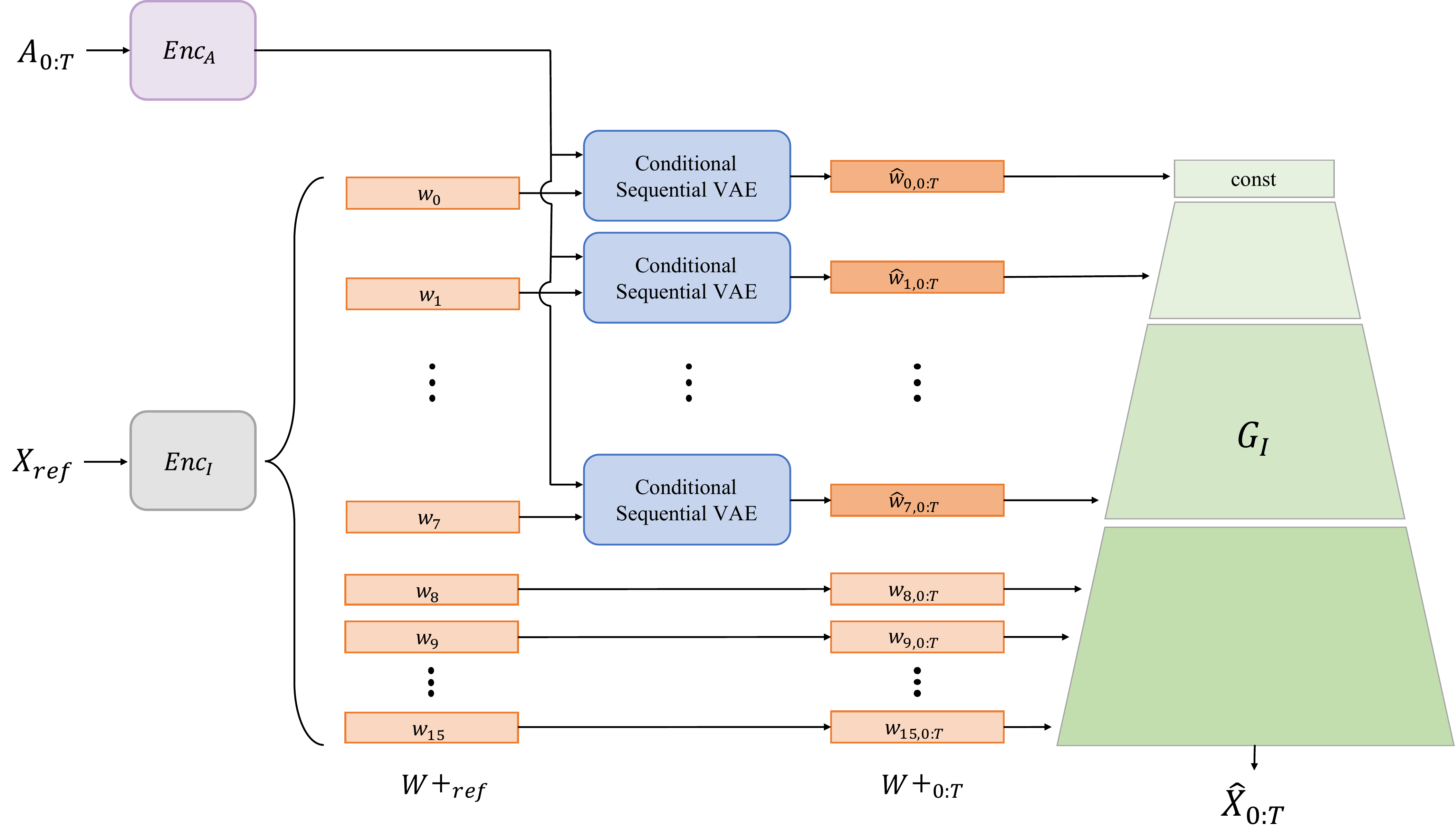}
    \caption{\small \textbf{Overall architecture.} }
    \label{fig:supple_fig1}
    \vspace{0.25in}
\end{figure}

\subsection{Posterior encoder and smoothing module}

\textbf{Posterior encoder:} Each style code, $w_{i, 0:T}$, in the latent style vector, $\mathbf{\mathcal{W}}+_{i, 0:T}$, is fed into the posterior encoder to output latent motion variables, $m_{i, 0:T}$. In detail, they first go through two fully-connected layers with LeakyReLU and Tanh activation functions. The hidden units and output units are 128 and 32. Then, they are concatenated with audio features, $a_{0:T}$, which have 32 hidden units through a fully-connected layer, and go into a LSTM layer followed by a fully-connected layer, which predicts means and standard deviations, $\mu_{0:T}$ and ${\sigma_{0:T}}$. Eventually, the motion latent variables, $m_{i, 0:T}$, can be sampled from the predicted means and standard deviations. \textbf{Smoothing module:} After the motion latent variables, $m_{i, 0:T}$, are concatenated with the audio features, $a_{0:T}$, they are fed into the smoothing module. The smoothing module consists of a 1D GaussianSmoothing filter, 1D convolution layer and LeakyReLU activation. All hidden units in the smoothing modules are 128. Then, the smoothing module outputs smoothed vectors, $c_{i, 0:T}$, following Eq.~\ref{eq:12}.

\begin{figure}[htbp]
    \centering
    \includegraphics[width=0.9\linewidth]{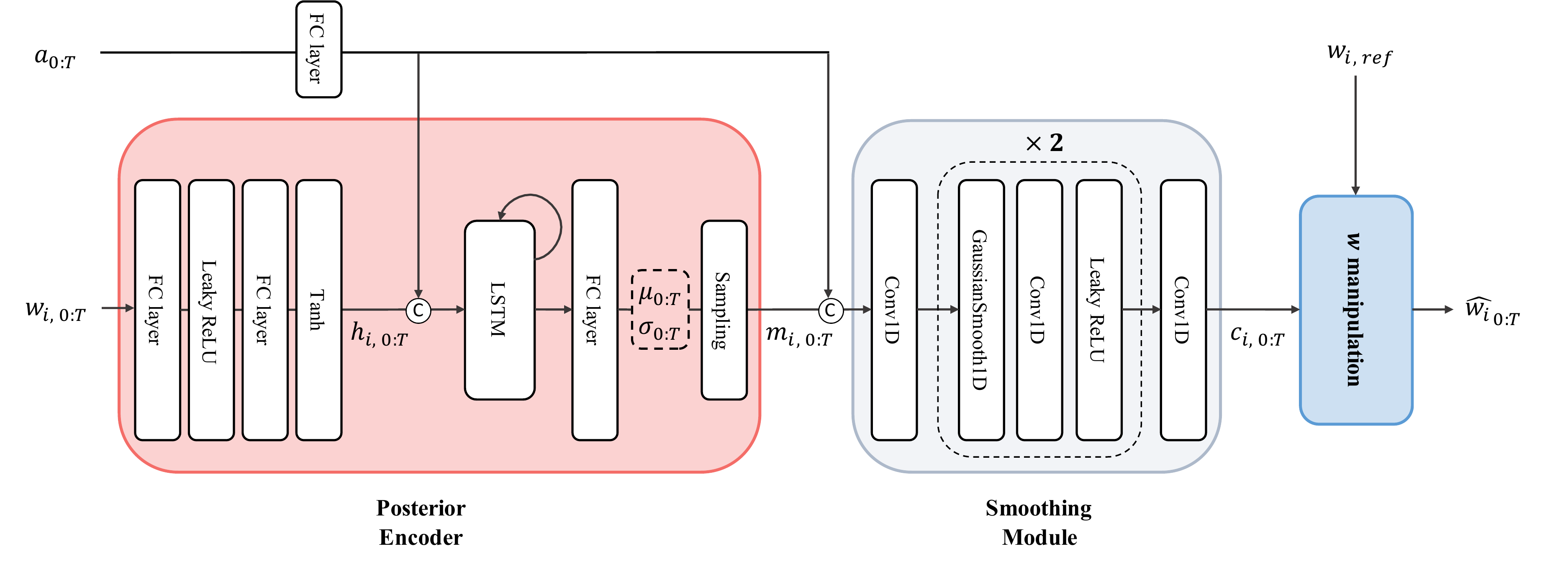}
    \caption{\small \textbf{Posterior encoder (red) and smoothing module (gray).}}
    \label{fig:supple_fig2} 
\end{figure}

\subsection{Prior encoder and normalizing flow}
\textbf{Prior encoder:} The prior encoder consists of two LSTM layers. The first LSTM layer takes the $h_{i,0}$, which is extracted from the reference image, as an input and outputs a latent variable, $m_{i,0}$. The second LSTM layer takes the previous latent variables, $m_{i,t-1}$ concatenated with the audio features, $a_{i,t}$ as an input and outputs the current latent variable, $m_{i,t}$. The both LSTM layers have 256 hidden units. The audio features have 32 hidden units through a fully-connected layer. \textbf{Normalizing flow:} Normalizing flow module consists of actnorm and affine coupling layers. The affine coupling layer contains 2 stacks of LSTM layers, each of which has 256 hidden units, to perform autoregressive causal transformation. Furthermore, we set the normalizing flow to be volume preserving~\cite{GIN}.

\begin{figure}[h]
    \vspace{0.25in}
    \centering
    \includegraphics[width=0.6\linewidth]{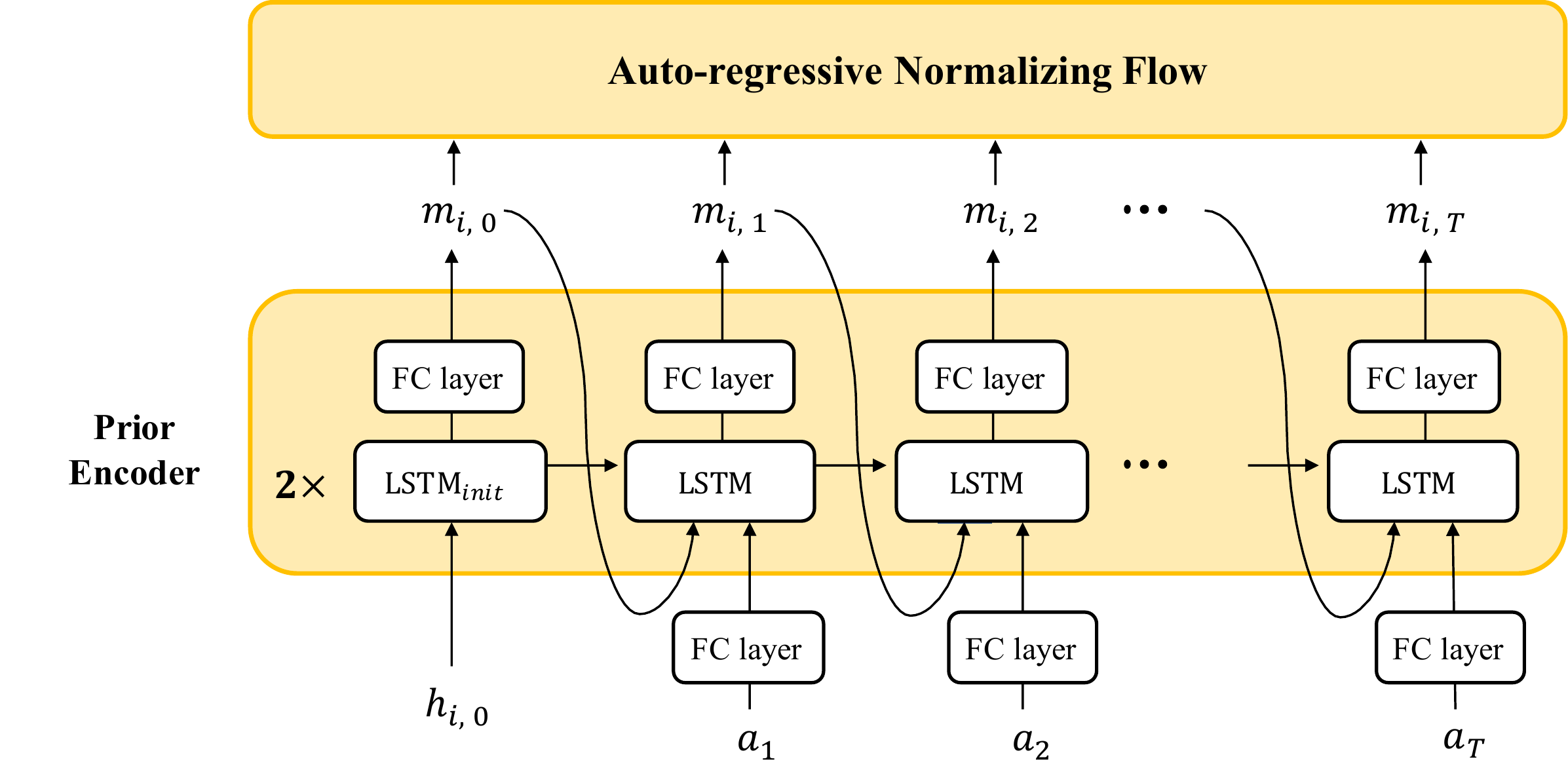}
    \caption{\small \textbf{Prior encoder and normalizing flow.}}
    \label{fig:my_label2}
\end{figure}

\section{Image encoder training}
\noindent \textbf{Dataset}
Image encoder $Enc_I$ is also trained on Voxceleb2~\cite{vox2} dataset. Description for Voxceleb2 is in  Sec.~\ref{dataset}. \\

\noindent \textbf{Model Architecture}
We adopt pixel2style2pixel~\cite{psp} (pSp) as the image encoder but perform some modifications in order to replace the StyleGAN2~\cite{stylegan2} with an alias-free StyleGAN generator~\cite{stylegan3}. An alias-free StyleGAN is aligned with previous versions of StyleGAN~\cite{stylegan,stylegan2}, but proposes better results in video generation by eliminating the alias through its improved architectural design.

In detail, following the idea of pSp, we adopt $\mathcal{W}+$ space where a style vector consist of $16$ style codes $w^{0,1, ..., 15}\in \mathcal{R}^{512}$. With the backbone network, which is a feature pyramid network~\cite{lin2017feature}, style codes are computed with three levels of feature maps and style mapping networks. $w^{0,1,2}$ are generated from small feature map, $w^{3,4,5,6}$ from the the medium feature map, and $w^{7,...,15}$ from the largest feature map. Specifically, the backbone network is an improved version of ResNet50~\cite{deng2019arcface} and the style mapping network is set of 2-strided convolutions followed by LeakyReLU activations~\cite{psp}.\\

\noindent \textbf{Training}
We follow the same training scheme proposed in Richardson et al.~\cite{psp}. Specifically, the loss is defined as follows:
\begin{equation}
    L_{pSp} = \lambda_1 L_2(x) + \lambda_2 L_{LPIPS}(x) + \lambda_3 L_{id} (x) + \lambda_4 L_{reg} (x)
\end{equation}
We set $\lambda_1, \lambda_2, \lambda_3, \lambda_4$ as $1.0, 0.8, 0.1, 0.01$, respectively. For more details, please refer to Richardson et al.~\cite{psp}.

\end{document}